\documentclass{bmvc2k}

\title{End-to-End Object Detection with\\ Adaptive Clustering Transformer}

\addauthor{Minghang Zheng}{minghang@pku.edu.cn}{1}
\addauthor{Peng Gao}{1155102382@link.cuhk.edu.hk}{2}
\addauthor{Renrui Zhang}{1700012927@pku.edu.cn}{2}
\addauthor{Kunchang Li}{kc.li@siat.ac.cn}{2}
\addauthor{Xiaogang Wang}{xgwang@ee.cuhk.edu.hk}{3}
\addauthor{Hongsheng Li}{hsli@ee.cuhk.edu.hk}{3}
\addauthor{Hao Dong}{hao.dong@pku.edu.cn}{1}

\addinstitution{CFCS, CS Dept., Peking University}
\addinstitution{Shanghai AI Laboratory}
\addinstitution{CUHK-SenseTime Joint Laboratory, \\The Chinese University of Hong Kong}

\runninghead{Zheng \etal}{ADAPTIVE CLUSTERING TRANSFORMER}

\def\etal{\emph{et al}\bmvaOneDot}

\usepackage{times}
\usepackage{epsfig}
\usepackage{graphicx}
\usepackage{amsmath}
\usepackage{amssymb}
\usepackage{booktabs}
\usepackage{multirow}
\usepackage{dcolumn}
\usepackage{wrapfig}
\usepackage{makecell}

\begin{document}

\maketitle

\begin{abstract}
End-to-end Object Detection with Transformer (DETR) performs object detection with Transformer and achieves comparable performance with two-stage object detection like Faster-RCNN. However, DETR needs huge computational resources for training and inference due to the high-resolution spatial inputs. In this paper, a novel variant of transformer named Adaptive Clustering Transformer (ACT) has been proposed to reduce the computation cost for high-resolution input. ACT clusters the query features \textbf{adaptively} using Locality Sensitive Hashing (LSH) and approximates the query-key interaction using the prototype-key interaction. ACT can reduce the quadratic O($N^2$) complexity inside self-attention into O($NK$) where K is the number of prototypes in each layer. ACT can be a drop-in module replacing the original self-attention module \textbf{without any training}. ACT achieves a good balance between accuracy and computation cost (FLOPs). The code is available as supplementary for the ease of experiment replication and verification. Code is released at \url{https://github.com/gaopengcuhk/SMCA-DETR/}
\end{abstract}

%-------------------------------------------------------------------------

\begin{figure}
    \centering
    \scalebox{0.95}[0.85]{
    \includegraphics[width=\linewidth]{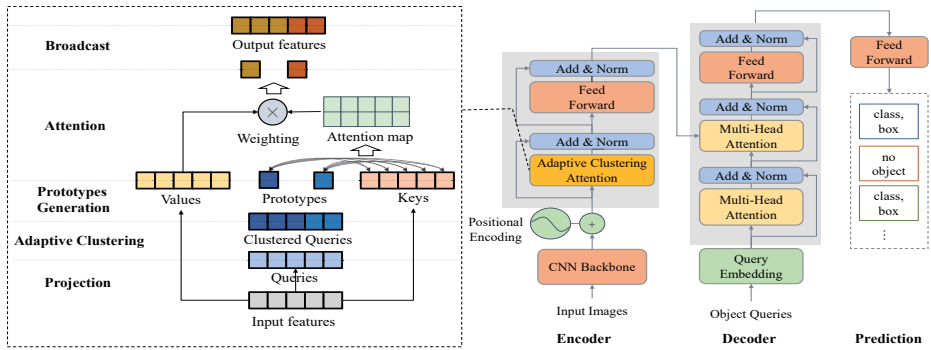}}
    \caption{The illustration of our adaptive clustering transformer. We use a small number of prototypes to represent the queries and only the attention map between prototypes and keys will be calculated. The number of prototypes will be automatically determined based on the distribution of queries. Finally, the attention output will be broadcast to the queries represented by the prototype.}
    \label{fig:adaptive_transformer}
\end{figure}
\section{Introduction}
\label{sec:intro}

Object detection is the task of predicting a set of bounding boxes and category labels for each predetermined object. Recently popular models~\cite{viola2001rapid,girshick2015fast,girshick2014rich,ren2015faster,redmon2016you,liu2016ssd} solve this task by generating a large number of regional proposals, predicting each proposal, and applying a non-maximum suppression procedure to eliminate those highly overlapping proposals. Two-stage object detection is difficult to deploy and debug due to the complex computation pipeline. 

Carion \etal~\cite{carion2020end} proposes a new method, called Detection Transformer or DETR, which uses an encoder-decoder transformer~\cite{vaswani2017attention} framework to solve this task in an intuitive way utilizing set prediction which has been explored in ~\cite{kuhn1955hungarian,stewart2016end,romera2016recurrent}.
% The main architecture is depicted in Figure~\ref{fig:detr}. 
%DETR uses a deep residual network (ResNet)~\cite{he2016deep} backbone to extract features and supplements it with a positional encoding before passing it into a transformer encoder. A transformer decoder then takes a small fixed number of learned positional embedding as object queries and additionally attends to the final layer of encoder iteratively. Finally, DETR passes the normalized output of the decoder layer to a feed-forward network (FFN) that predicts either a detection (class and bounding box) or a ``no object'' class. DETR is trained with Hungarian loss~\cite{kuhn1955hungarian} by matching predicted bounding boxes with grounding truth annotations.
Thanks to the powerful learning ability of Transformer~\cite{vaswani2017attention}, DETR can perform set prediction end-to-end without resorting to human-designed prior like anchor and region proposal thus resulting in a much simpler object detection framework. However, DETR suffers from high computational complexity in the encoder. To achieve good performance, DETR needs a high-resolution image which will increase the computation in the encoder quadratically due to the all-pairs interaction for all positions. 

Although many improvements of transformer~\cite{kitaev2020reformer,choromanski2020rethinking,chen2019graph,zhu2019asymmetric,goyal2020power,katharopoulos2020transformers,gao2019multi,chen20182, vyas2020fast} can reduce the computation complexity, variants of transformer change the architecture of transformer which require huge trial-and-error cost due to the slow convergence of DETR(1920 GPU hours for single V100). One natural question to ask is whether we can improve the performance and computational trade-off of DETR with acceptable computing resources?

We propose a novel Adaptive Clustering Transformer(ACT) which can serve as a drop-in module on the original DETR framework by replacing the transformer in the encoder. ACT is fully compatible with the original transformer and thus does not require retraining. The accuracy gap between ACT and the original transformer can be further closed by equipping with Multi-Task Knowledge Distillation(MTKD). MTKD can also enable seamless switch between models with different FLOPs and Accuracy during inference. 

Two observations of DETR motivate our design.

\begin{figure}
    \centering
    \scalebox{0.95}[0.85]{
    \begin{minipage}[]{0.48\textwidth}
        \centering
        \includegraphics[width=6cm]{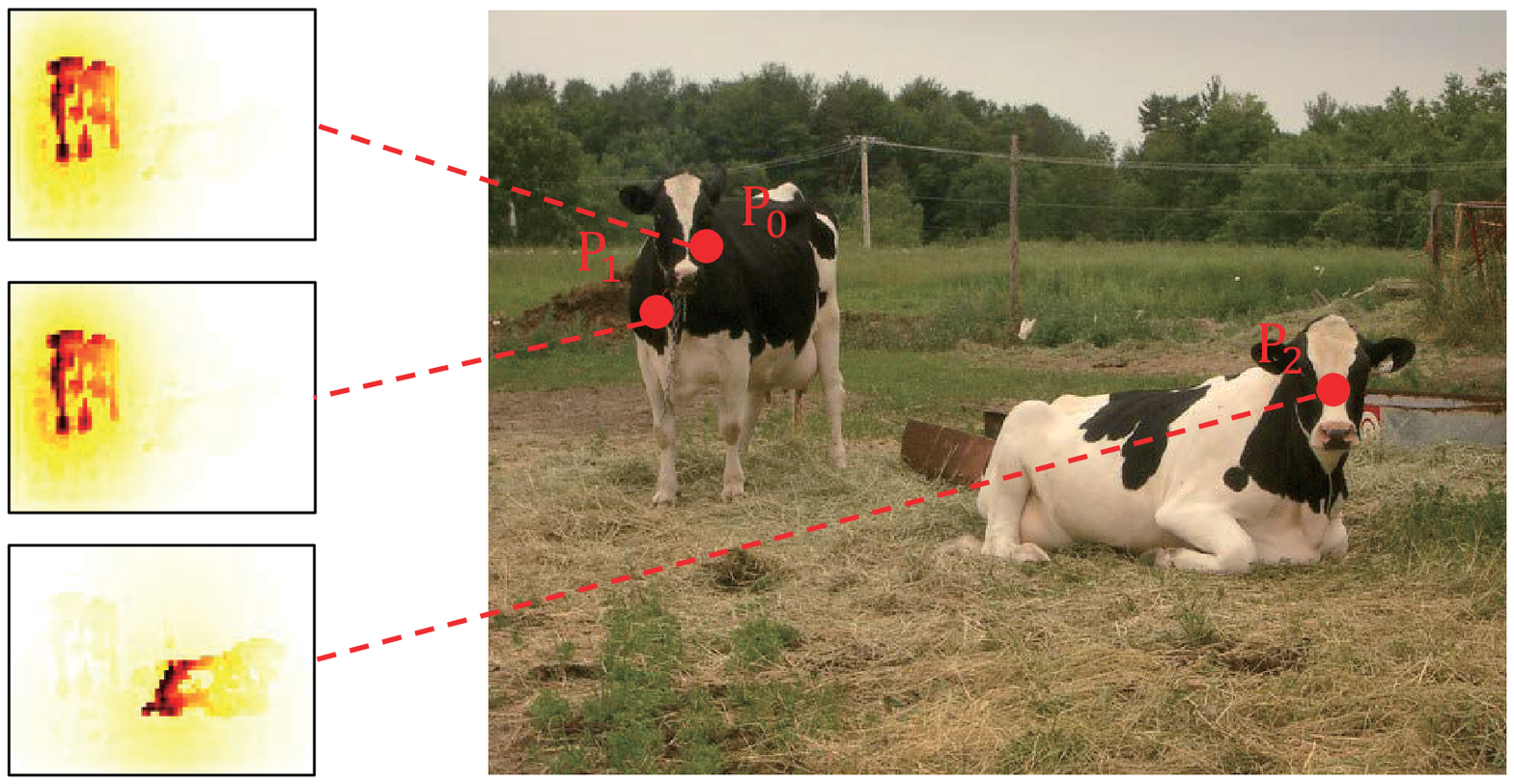}
        \caption{The attention map of some points in the last layer of the transformer encoder. The darker the color, the greater the weight.}
        \label{fig:weights}
    \end{minipage}
    \hspace{0.1in}
    \begin{minipage}[]{0.48\textwidth}
        \centering
        \includegraphics[width=6cm]{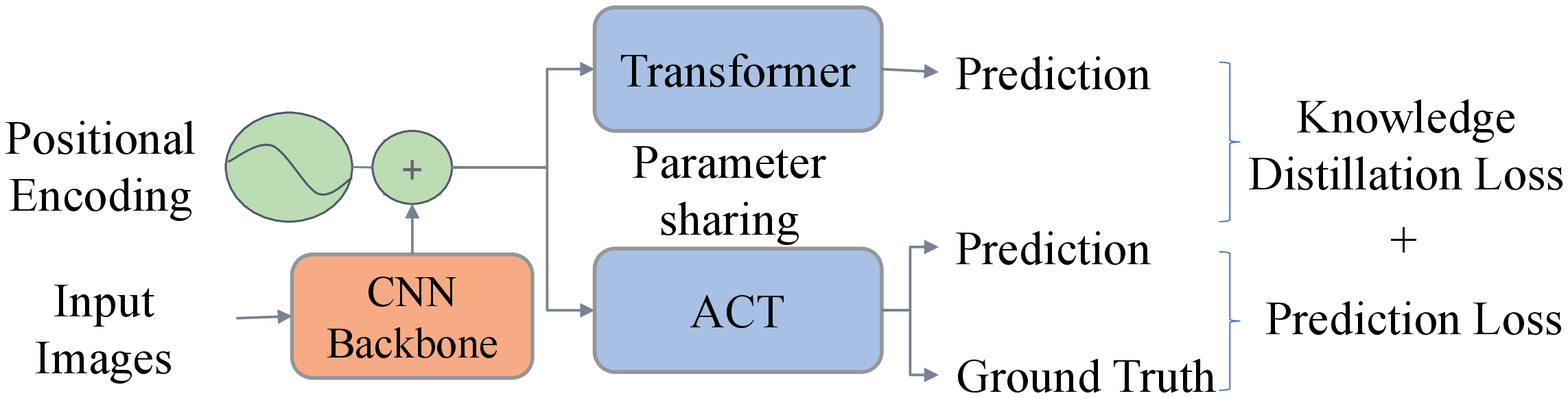}
        \caption{Multi-Task Knowledge Distillation. Image features will be extracted by the CNN backbone first. The extracted feature will be passed into ACT and the original transformer parallel. To enable a seamless switch between ACT and the orignal transformer, MTKD will guide the training.}
        \label{fig:kd}
    \end{minipage}}
\end{figure}

\textbf{Encoder Attention Redundancy} 
Inside the encoder of DETR, features at each position will collect information from other spatial positions adaptively using the attention mechanism. We observe that features that are semantically similar and spatially close to each other will generate similar attention maps and vice versa. As shown in Figure ~\ref{fig:weights}, the attention map for $P_0$ and $P_1$ are similar to each other and contain redundancy while distant points $P_0$ and $P_3$ demonstrate a completely different attention pattern. The redundancy in self-attention motivates ACT to choose representative prototypes and broadcast the feature update of prototypes to its nearest neighbor. 
% This inspires us to use some prototypes to represent those points where the attention is concentrated in the same place, that is, the points where the query have a relatively close distance. This can greatly reduce the calculations caused by redundancy, because those points with similar attention map will only be calculated once.

\textbf{Encoder Feature Diversity} 
We observe that as the encoder goes deeper, features will be similar as each feature will collect information from each other. 
%To verify this hypothesis, we calculated the average distance between the features in each layer to their center for the first 100 pictures on the training set. As shown in figure~\ref{fig:distance}, feature similarity will decrease as we go deeper which consolidates our hypothesis. 
Besides, for different inputs, the feature distribution in each encoder layer is quite different.
These observations motivates us to adaptively determine the number of prototypes based on the distribution of features among each layer instead of a static number of cluster centers. 

To solve the \textbf{Encoder Attention Redundancy}, ACT clusters similar query features together and only calculates the key-query attentions for representative prototypes according to the average of the features in the cluster. After calculating the features updates for prototypes, the updated features will be broadcast to its neighbors according to the euclidean distance on the query feature space. 
%A naive idea is to cluster query features using K-means with a pre-defined number of centers for all images. As shown in the experiment part, the K-means cluster will significantly deteriorate the performance of pre-trained DETR. 
\textbf{Encoder Feature Diversity} motivate us to design an adaptive clustering algorithm which can cluster features according to the distribution of feature for each input and each encoder layer. Thus we choose a multi-round Exact Euclidean Locality Sensitivity Hashing (E2LSH) which can perform query features distribution-aware clustering. 

Experiments show that we reduce the FLOPS of DETR from 73.4 Gflops to 58.2 Gflops (excluding Backbone Resnet FLOPs) \textbf{without any training process}, while the loss in AP is only 0.7\%. The loss in AP can be further reduced to 0.2\% by a Multi-Task Knowledge Distillation. 

Our main contributions are summarised below.
\begin{itemize}
\setlength{\parsep}{0pt}
\setlength{\parskip}{0pt}

\item We develop a novel method called Adaptive Clustering Transformer (ACT) which can reduce the inference cost of DETR. 
%The core idea of ACT is to select representative prototypes from queries using lightweight LSH and then broadcast the feature update of selected prototypes to its nearest query. 
ACT can reduce the quadratic complexity of the original transformer, at the same time ACT is fully compatible with the original transformer.
%ACT adaptively select protopyes from query using lightweight Locality Sensitivity Hashing (LSH).a small number of prototypes to represent the queries in order to reduce the complexity of attention calculation, while ensuring compatibility with the original transformer.

\item We reduce the FLOPS of DETR from 73.4 Gflops to 58.2 Gflops (excluding Backbone Resnet FLOPs) \textbf{without any training process}, while the loss in AP is only 0.7\%.

\item We have further reduced the loss in AP to 0.2\% through a Multi-Task Knowledge Distillation (MTKD) which enables a seamless switch between ACT and the original transformer.

\end{itemize}

\section{Related Work}
\subsection{Review of Attention Model on NLP and CV}
Attention model~\cite{bahdanau2014neural,xu2015show,jiang2019video} has been widely used in computer vision (CV) and natural language processing (NLP) fields due to the in-built adaptive information aggregation mechanism. We mainly focus on one branch of attention called transformer. Transformer performs information exchange between all pairs of entities like words in NLP or regions in CV. Transformer has achieved state-of-the-art performance on Machine Translation~\cite{vaswani2017attention}, Object Detection~\cite{carion2020end}, Multimodality Reasoning~\cite{gao2019dynamic,yu2019deep,nguyen2018improved}, Image Classfication~\cite{dosovitskiy2020image, liu2021swin}, Video Classfication~\cite{wang2018non} and Language Understanding~\cite{devlin2018bert}. While transformer has achieved good performance on different scenarios, but it is hard to scale due to the quadratic complexity with respect to the length of the input sequence. Many modifications of transformer have been proposed to tackle the computation bottleneck of transformer.

Reformer~\cite{kitaev2020reformer} proposed sharing key and query and use Locality Sensitivity Hashing (LSH)~\cite{datar2004locality} to cluster features near into one cluster, then perform information exchange inside each cluster. Performer~\cite{choromanski2020rethinking} approximates the softmax between key and query interaction using Positive Orthogonal Random Features (PORF) with provable approximate error with linear complexity. Linear Attention~\cite{katharopoulos2020transformers} utilizes association property to Key-Query-Value multiplication from quadratic complexity into linear complexity. Progressive Elimination~\cite{goyal2020power} finds that redundancy exists in transformer and progressively eliminates the input of each layer and achieves comparable performance with the original transformer by reducing computation cost. Asymmetric Attention~\cite{zhu2019asymmetric} summarises key features into a few key vectors using multi-scale pooling over key features thus reduce the computation complexity. Global Graph Reasoning~\cite{chen2019graph} transforms the original input into global vectors utilizing weighted pooling and then perform information exchange over the compact global vectors.  

Previously mentioned methods modified the structure of the original transformer and need huge resources for training and inference. Our proposed Adaptive Clustering Transformer (ACT) shares the same structure as the original transformer. ACT reduces the computation cost of transformer without re-training. Besides, the performance gap between ACT and the original transformer can be further reduced with a few epochs of fine-tuning knowledge distillation~\cite{hinton2015distilling,zhang2019your}.

\subsection{Object Detection using Deep Learning}
The main framework of object detection is dominated by performing classification over a sliding window. Viona-Jones Face detector~\cite{viola2001rapid} first introduce the idea of sliding window approach into face detection with adaboost~\cite{friedman2000additive}. After the successful application of CNN on object classification~\cite{krizhevsky2017imagenet,he2016deep}, deep features have been applied to object detection. Previous research of object detection using deep features can be divided into two-stage and one-stage object detection. RCNN, Fast RCNN and Faster RCNN~\cite{girshick2014rich,girshick2015fast,ren2015faster} are two-stage solution while YOLO~\cite{redmon2016you} and SSD~\cite{liu2016ssd} are one-stage solution. Previous methods on object detection are suffered by  complex post-processing pipeline (NMS)~\cite{neubeck2006efficient,bodla2017soft}, imbalanced loss~\cite{lin2017focal}, and hand-crafted anchor~\cite{liu2016ssd,redmon2016you} which increase the difficulty of training and deployment. Unlike sliding-window approaches, object detection has been formulated as a permutation-invariant set prediction problem. Steward et al~\cite{stewart2016end} proposed an end-to-end people detection which encodes image feature using CNN and decodes the bounding box sequentially using LSTM~\cite{hochreiter1997long}. The predicted bounding box will be matched with ground truth using Hungarian loss~\cite{kuhn1955hungarian} and trained end-to-end. Recurrent instance segmentation~\cite{romera2016recurrent} adds an extra segmentation head over the end-to-end object detection framework and successfully tests the idea on instance segmentation. Recently DETR~\cite{carion2020end} has successfully made the performance of set-prediction approaches comparable with two-stage Faster RCNN approaches by replacing LSTM~\cite{hochreiter1997long}with much powerful Transformer~\cite{vaswani2017attention}. End-to-end set prediction problem significantly simplified the pipeline of object detection and reduce the need for hand-crafted prior. However, the convergence of end-to-end set prediction is slow and need huge inference cost caused by the quadratic complexity of self-attention. Some variants of DETR have also been proposed to solve the problems. Deformable DETR~\cite{zhu2020deformable} accelerates the convergence speed via learnable sparse sampling coupled with multi-scale deformable encoder. SMCA~\cite{gao2021fast} introduces Gaussian prior in the transformer decoder and achieved an increase in convergence speed. Different from Deformable DETR and SMCA, whose improvement is mainly on training process, our proposed ACT target to reduce the inference computation cost of DETR without the need for retraining. 

%\section{Adaptive Clustering Transformer}
%\subsection{Detr for Object Detection}

\section{Adaptive Clustering Transformer}

\subsection{Main Structure of DETR}
Figure~\ref{fig:adaptive_transformer} also shows the three stages of DETR. In the encoder, an ImageNet-pre-trained ResNet model is used to extract 2D features from the input image. The positional encoding module uses sine and cosine functions with different frequencies to encode spatial information. DETR flattens the 2D features and supplements them with the positional encoding and passes them to the 6-layer transformer encoder. Each layer of the encoder has the same structure, including an 8-head self-attention module and an FFN module. The decoder then takes as input a small fixed number of learned positional embeddings, which are called object queries, and additionally attends to the encoder output. The decoder also has 6 layers, and each layer contains an 8-head self-attention module, an 8-head co-attention module, and an FFN module. Finally, DETR passes each output of the decoder to a shared feed-forward network that predicts either a detection (class and bounding box) or a ``no object" class.

\label{section:review}

\subsection{Adaptive Clustering Transformer}
\textbf{Determine Prototypes} 
%\subsection{Determine the Prototypes}
We use Locality Sensitivity Hashing (LSH) to adaptively aggregate those queries with a small Euclidean distance. LSH is a powerful tool to solve the Nearest Neighbour Search problem. We call a hashing scheme locality-sensitive if nearby vectors get the same hash with high probability and distant ones do not. By controlling the parameters of the hash function and the number of hashing rounds, we let all vectors with a distance less than $\epsilon$ fall into the same hash bucket with a probability greater than $p$.

We choose Exact Euclidean Locality Sensitive Hashing (E2LSH)~\cite{datar2004locality} as our hash function:
\begin{equation}
    h(\vec{v})=\lfloor \frac{\vec{a} \cdot \vec{v} + b}{r} \rfloor
    \label{con:lsh}
\end{equation}
where $h:\mathbb{R}^d \rightarrow \mathbb{Z}$ is the hash function, $r$ is a hyper-parameter, $\vec{a}, b$ are random variables satisfying $\vec{a}=(a_1,a_2,...,a_d)$ with $a_i\sim \mathcal{N}~(0,1)$ and $b\sim\mathcal{U}~(0,r)$. We will apply $L$ rounds of LSH to increase the credibility of the results. The final hash value will be obtained by equation~\ref{con:multi_rounds}.
\begin{equation}
    h(\vec{v})=\sum_{i=0}^{L-1}B^ih_i(\vec{v})
    \label{con:multi_rounds}
\end{equation}
where each $h_i$ is obtained by equation~\ref{con:lsh} with independently sampled parameters $\vec{a}$ and $b$, and $B$ is a constant equal to 4 in our experiments.

%Figure~\ref{fig:lsh} shows the principle of our hash function. 
Each hash function $h_i$ can be regarded as a set of parallel hyperplanes with random normal vector $\vec{a_i}$ and offset $b_i$. The hyper-parameter $r$ controls the spacing of the hyperplanes. The greater the $r$, the greater the spacing. Furthermore, $L$ hash functions divide the space into several cells, and the vectors falling into the same cell will obtain the same hash value. Obviously, the closer the Euclidean distance, the greater the probability that the vectors fall into the same cell.

% \begin{figure}
%     \begin{center}
%         \includegraphics[width=\linewidth]{images/e2lsh.eps}
%     \end{center}
%     \caption{The principle of our hash function. Each hash function $h_i$ can be regarded as a set of parallel hyperplanes with random normal vector $\vec{a_i}$ and offset $b_i$. The hyper-parameter $r$ controls the spacing of the hyperplanes. $L$ hash functions divides the space into several cells. The vectors falling into the same cell will obtain the same hash value.}
%     \label{fig:lsh}
% \end{figure}

% \begin{figure}
%     \centering
%     \begin{minipage}[t]{0.48\textwidth}
%         \centering
%         \includegraphics[width=6cm]{images/e2lsh.eps}
%         \caption{The principle of our hash function. Each hash function $h_i$ can be regarded as a set of parallel hyperplanes with random normal vector $\vec{a_i}$ and offset $b_i$. The hyper-parameter $r$ controls the spacing of the hyperplanes. $L$ hash functions divides the space into several cells. The vectors falling into the same cell will obtain the same hash value.}
%         \label{fig:lsh}
%     \end{minipage}
%     \begin{minipage}[t]{0.48\textwidth}
%         \centering
%         \includegraphics[width=6cm]{images/kd.eps}
%         \caption{Multi-Task Knowledge Distillation.  Image features will be extracted by the pre-trained CNN backbone first. The extracted feature will be passed into ACT and the original transformer parallel. To enable seamless switch between ACT and orignal transformer, MTKD will guide the training.}
%         \label{fig:kd}
%     \end{minipage}
% \end{figure}

To obtain the prototypes, we calculate the hash value for each query firstly. Then, queries with the same hash value will be grouped into one cluster, and the prototype of this cluster is the center of these queries. More formally, we define $Q\in \mathbb{R}^{N\times D_k}$ as the queries and $P\in \mathbb{R}^{C\times D_k}$ as the prototypes, where $C$ is the number of clusters. Let $G_i$ represent the index of the cluster that $Q_i$ belongs to. The prototype of the $j-th$ cluster can be obtained by equation~\ref{con:prototypes}.
\begin{equation}
    P_j = \frac{\sum_{i, G_i=j}Q_i}{\sum_{i, G_i=j}1}
    \label{con:prototypes}
\end{equation}

%\label{section:prototypes}

\textbf{Estimate Attention Output} 
%\subsection{Estimate Attention Output}
After the previous step, each group of queries is represented by a prototype. Thus, only the attention map between prototypes and keys need to be calculated. Then, we get the target vector for each prototype and broadcast it to each original query. Thus, we get an estimation of the attention output. Compared with the exact attention calculation, we reduce the complexity from $O(NMD_k+NMD_v)$ to $O(NLD_K+CMD_k+CMD_v)$, where $C$ is the number of prototypes and $L$ is the number of hash rounds, both of which are much smaller than $N$ and $M$.

More formally, we define $K\in \mathbb{R}^{M\times D_k}$ as the keys and $V\in \mathbb{R}^{C\times D_v}$ as the values. We get the estimate of attention output $\Tilde{V^o}$ by the following equations:
\begin{gather}
    \Tilde{A} = softmax(PK^T/\sqrt{D_k})\\
    \Tilde{W} = \Tilde{A}V\\
    \Tilde{V^o_i} = \Tilde{W_j}\ ,\ if\ G_i=j
\end{gather}
where the softmax function is applied row-wise and $G_i$ represents the index of the cluster that $Q_i$ belongs to.
\label{section:act}

\subsection{Multi-Task Knowledge Distillation}
Although ACT can reduce the computation complexity of DETR without retraining, we show that Multi-Task Knowledge Distillation(MTKD) can further improve ACT with a few-epoch of fine-tuning and produce a better balance between performance and computation. The pipeline of MTKD is illustrated in figure ~\ref{fig:kd}. Image features will be extracted by the pre-trained CNN backbone first. The extracted feature will be passed into ACT and the original transformer parallelly. To enable a seamless switch between ACT and the original transformer, MTKD will guide the training. The training loss is denoted below:
\begin{equation}
    L = L_{pred}(Y,Y_2)  + L_{KD}(B_1,B_2.detach())
\end{equation}
where $Y$ represents the ground truth, $B_1$ represents the predicted bounding box of ACT, and $B_2, Y_2$ represent the predicted bounding box and full prediction of DETR. $L_{pred}(Y,Y_2)$ is the original loss between the ground truth and the prediction of DETR. $L_{KD}(B_1, B_2)$ is the knowledge distillation loss which minimizes the L2 distance between the predicted bounding box of ACT and DETR.

The training loss aims to train the original transformer jointly with knowledge transfer between full prediction and approximated prediction which enables a seamless switch between ACT and Transformer. The knowledge transformer includes region classification and regression distillation. The regression branch is more sensitive to the approximated error introduced by ACT than the classification branch. Thus, we only transfer the knowledge of the bounding box regression branch. We observe much faster convergence by transferring the box regression branch only.  
\label{section:distillation}

\begin{figure}
    \centering
    \scalebox{0.9}{
    \begin{minipage}[t]{0.48\textwidth}
        \centering
        \includegraphics[width=6.2cm]{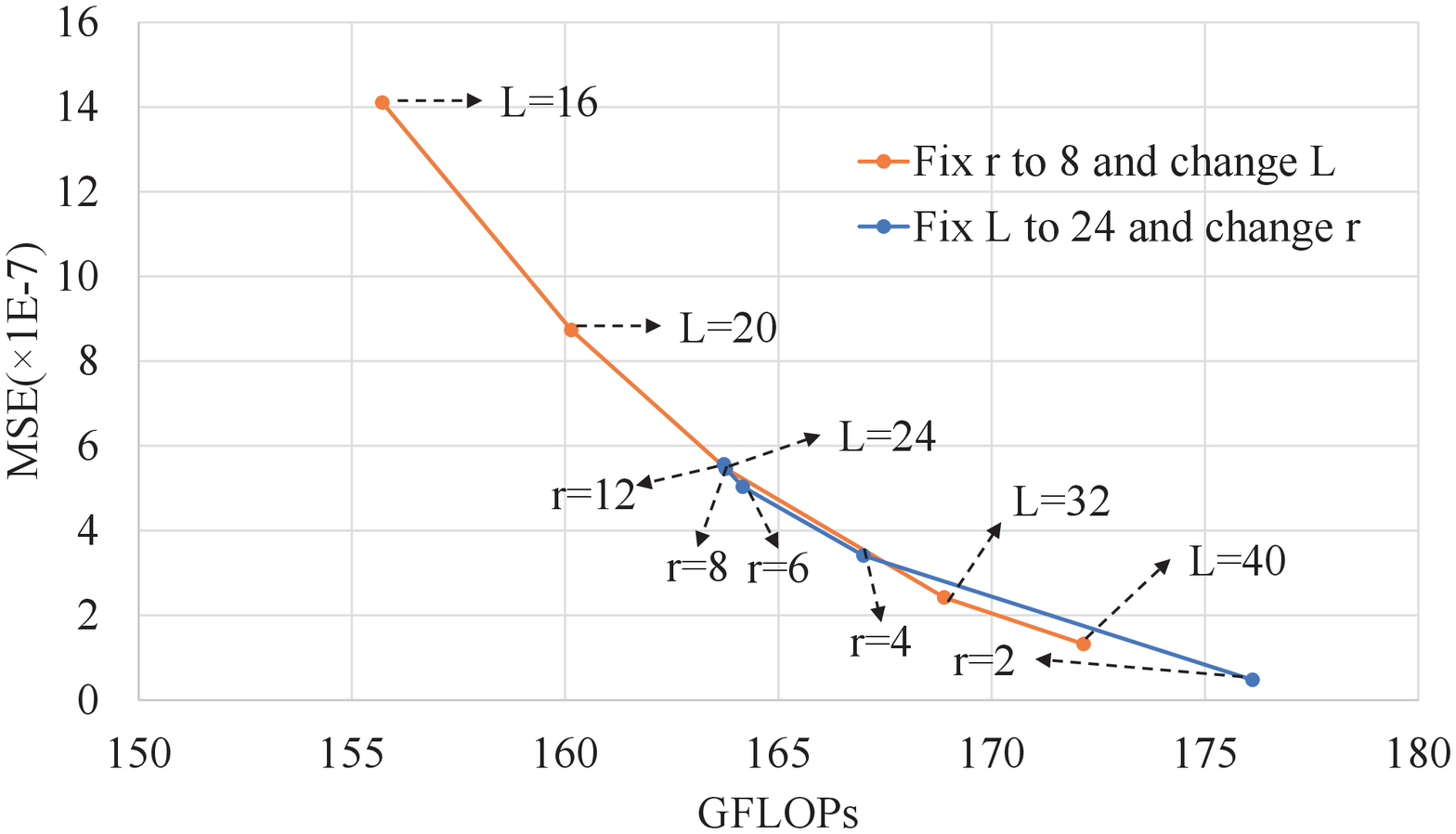}
        \caption{The mean square error between the estimated attention map and the true attention map under a certain computational budget. We fix $L$ to 24 and set $r$ to 2,4,6,8,12 respectively. Then we fix $r$ to 8 and set $L$ to 16, 20, 24, 32 respectively.}
        \label{fig:parameters}
    \end{minipage}
    \hspace{0.1in}
    \begin{minipage}[t]{0.48\textwidth}
        \centering
        \includegraphics[width=6cm]{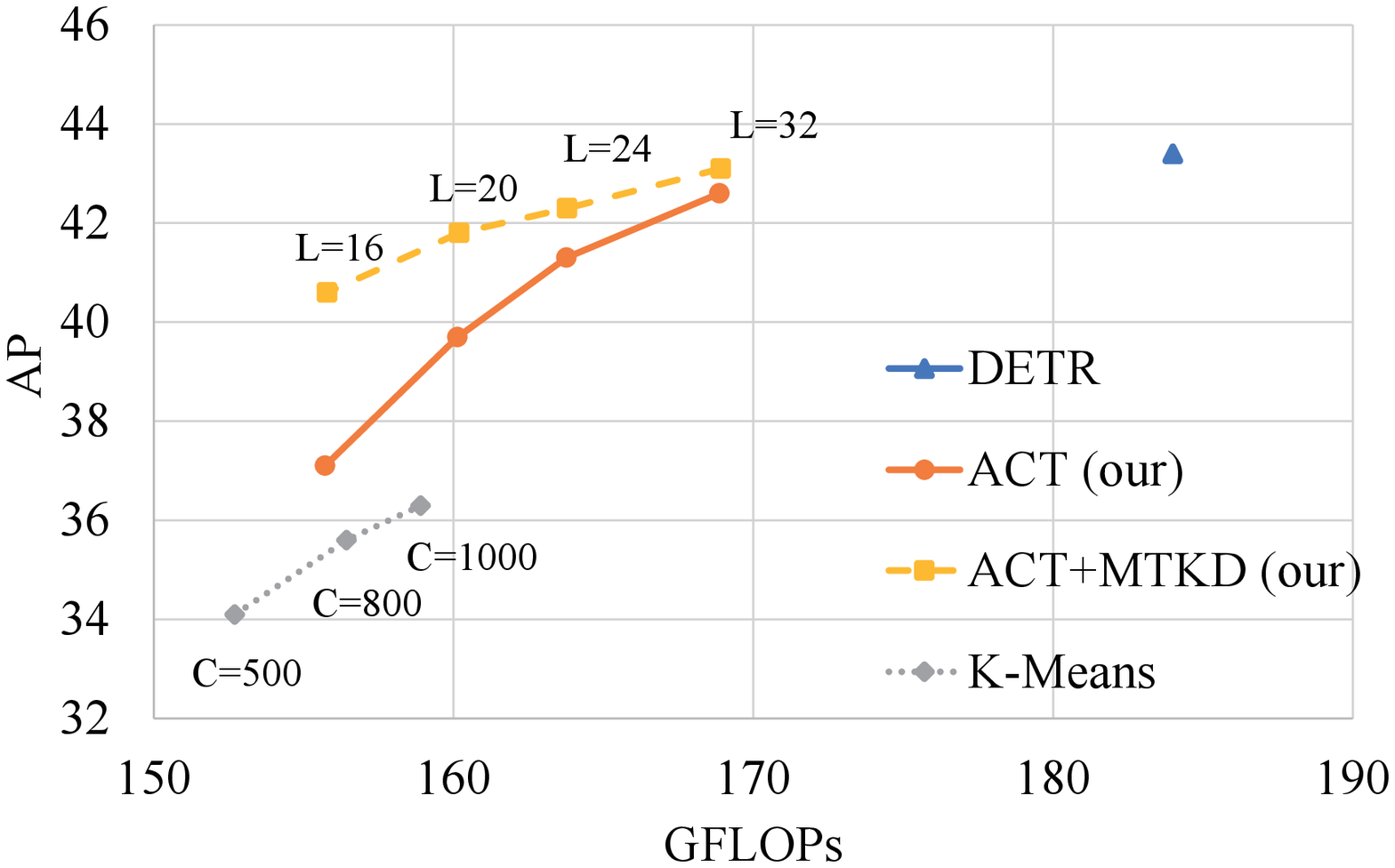}
        \caption{We compare the AP of ACT with DETR-DC5 and the K-mean clustering. We refer to hash rounds in our model as $L$ and refer to the number of clusters in K-means as $C$.}
        \label{fig:results}
    \end{minipage}}
\end{figure}

\section{Experiment}
\subsection{Dataset}
We perform experiments on COCO 2017 detection dataset~\cite{lin2014microsoft}, which containing 118k training images and 5k validation
images. Each image in the dataset contains up to 63 instances of different sizes. We report AP as bbox AP, the integral metric over multiple thresholds. We also report the average FLOPs for the first 100 images in the COCO 2017 validation set. Only the FLOPs of convolutional layers, fully connected layers, matrix operations in attention, E2LSH, and clustering will be considered.
\subsection{Experiment Setup}
We choose the pre-trained DETR-DC5 model~\cite{carion2020end} as our baseline. It uses the deep residual network~\cite{he2016deep} with 50 layers (ResNet-50) as the backbone and increases the feature resolution by adding a dilated convolution~\cite{yu2015multi} to the last stage of the backbone and removing a stride from the first convolution of this stage. DETR-DC5 contains 6 encoder layers and 6 decoder layers with 8 attention heads.

We replace the attention module in the encoder with our adaptive clustering attention while keeping the other parts unchanged.  We randomly sample 1000 images on the training set and calculate the mean square error between our estimated attention map and the true attention map to determine an appropriate hyper-parameter $r$ and control the FLOPs of the model by changing the hyper-parameter $L$, where $L$ represent the round of E2LSH and $r$ represent the interval of the hashing hyperplanes.

During inference, we resize the input images such that the shortest side is at most 800 pixels while the longest at most 1333. For Multi-Task Knowledge Distillation(MTKD), we adopt a random crop augmentation which is also used in DETR. We use a simple L2 norm for regression distillation and the weight of KD loss is set to 1. MTKD performs fine-tuning over the pre-trained model for 5 epochs with the learning rate of $10^{-5}$ and continues running for 2 epochs by reducing the learning rate by 1/10. MTKD is optimised by AdamW~\cite{kingma2014adam,loshchilov2018fixing}.

% \begin{figure}
%     \centering
%     \includegraphics[width=\linewidth]{images/parameter.eps}
%     \caption{The mean square error between the estimated attention map and the true attention map under an certain computational budget. We fix $L$ to 24 and set $r$ to 2,4,6,8,12 respectively. Then we fix $r$ to 8 and set $L$ to 16, 20, 24, 32 respectively.}
%     \label{fig:parameters}
% \end{figure}

\subsection{Ablation Study}
The hyper-parameters of the E2LSH have a great influence on the quality of the approximation and the FLOPs of the model. In our ablation analysis, we explore how these hyperparameters affect the results and try to determine appropriate hyper-parameters.

We randomly sample 1000 images on the training set and input the images into DETR and our ACT respectively to obtain the attention map in different encoder layers and different attention heads. We calculate the mean square error between our estimated attention map and the true attention map. We perform two sets of experiments. First, we fix $L$ to 24 and set $r$ to 2,4,6,8,12 respectively. Then, we fix $r$ to 8 and set $L$ to 16, 20, 24, 32 respectively. 
%The mean square error and FLOPs are listed in Table~\ref{tab:parameters}.

The results are shown in Figure~\ref{fig:parameters}. Firstly, the estimation error decreases with an increase of $L$ and a decrease of $r$. Secondly, when $r$ is greater than 6, continuing to increase $r$ has little effect on estimation errors and FLOPs. Therefore, it is a better choice to obtain models of different FLOPs by the change of $L$. Finally, we found that when $r$ is less than or equal to 6, continuing to reduce $r$ will cause a larger increase in FLOPs while a smaller decrease in error, which is not cost-effective. Thus, in all subsequent experiments, we fix $r$ to 8. 

Another significant discovery is that adaptively clustering keys using our method can also achieve a good result. We perform experiments on clustering queries, clustering keys, and clustering both queries and keys respectively. We adjust the hyper-parameter $L$ to ensure that the three experiments have similar FLOPs, and we compare the AP on the validation set. The results are given in Table~\ref{tab:cluster}. As we can see, these three methods have achieved similar AP under the same FLOPs, which also proves the generalization of our model. This means that for some models where the number of keys is significantly more than the number of queries, we can adaptively cluster the keys to obtain higher efficiency.

% \begin{table} 
%     \centering
%     \caption{The mean square error and the FLOPs of our model with different hyperparameters $L$ and $r$. Both are the lower the better.}
%     \begin{tabular}{cccc}
%         \toprule
%         $L$&$r$& GFLOPs & MSE($\times10^{-7}$)\\
%         \midrule
%         \multirow{5}*{24}& 2& 176.12&0.475\\
%         & 4&167.00 &3.41\\
%         & 6&164.17 &5.03\\
%         & 8&163.77 &5.46\\
%         & 12&163.63 &5.56\\
%         \midrule
%         16 & \multirow{4}*{8}& 155.72& 14.1\\
%         20 & &160.15 &8.73 \\
%         24 & & 163.77&5.46 \\
%         32 & & 168.89&2.42 \\
%         \bottomrule
%     \end{tabular}
    
%     \label{tab:parameters}
% \end{table}

% \begin{figure}
%     \centering
%     \includegraphics[width=\linewidth]{images/result.eps}
%     \caption{we compare the AP of ACT with DETR-DC5 and the K-mean clustering. We refer to hash rounds in our model as $L$ and refer to the number of clusters in K-means as $C$.}
%     \label{fig:results}
% \end{figure}

\begin{table}
    \centering
    \renewcommand\tabcolsep{12.0pt}
    \scalebox{0.9}[0.8]{
    \begin{minipage}[]{\textwidth}
    \begin{tabular}{cccccc}
        \toprule
         Model&GFLOPs&AP&AP$_L$&AP$_M$&AP$_S$  \\
         \midrule
         Backbone (ResNet50-DC5) &110.7&&&& \\
         \midrule
         DETR-DC5~\cite{carion2020end}& +73.4& \textbf{43.3} & \textbf{61.1}&\textbf{47.3}&22.5 \\
         Faster RCNN-DC5~\cite{ren2015faster}& +209.3 & 41.1 & 55.0 & 45.9 & \textbf{22.9}\\
        %  Deformable DETR~\cite{zhu2020deformable}& 173& \textbf{43.8} & 58.0 & 47.1 & \textbf{26.4} \\
         \midrule
         ACT (L=32)& +58.2& \textbf{42.6} & \textbf{61.1}&\textbf{46.8}&\textbf{21.4} \\
         ACT (L=24)& +53.1& 41.3 & 60.6&45.9&19.2 \\
         ACT (L=20)& +49.4& 39.7 & 60.3&44.2&16.9 \\
         ACT (L=16)& +45.0& 37.1 & 58.8&41.3&13.9 \\
         \midrule
         ACT+MTKD (L=32)& +58.2& \textbf{43.1} & \textbf{61.4}&\textbf{47.1}&\textbf{22.2} \\
         ACT+MTKD (L=24)& +53.1& 42.3 & 61.0&46.4&21.3 \\
         ACT+MTKD (L=20)& +49.5& 41.8 & 60.7&45.6&20.6 \\
         ACT+MTKD (L=16)& +45.1& 40.6 & 59.7&44.3&18.5 \\
         \bottomrule
    \end{tabular}
    \caption{We compare the AP of our model with DETR-DC5 and Faster RCNN in detail. DETR-DC5 and Faster RCNN use dilated ResNet-50 as the backbone. The sign `+' in GFLOPs column refers to the flops increased relative to the backbone. We refer to the bbox AP of the large, medium, and large size instances as AP$_L$, AP$_M$, and AP$_S$ respectively.}
    \end{minipage}}
    \label{tab:ap}
\end{table}

\begin{table} 
    \centering
    \scalebox{0.9}[0.8]{
    \begin{minipage}[]{0.47\textwidth}
        \centering
        \renewcommand\tabcolsep{8.0pt}
        %\scalebox{0.9}[0.85]{
        \begin{tabular}{cccc}
            \toprule
             \makecell{Cluster \\ queries}&\makecell{Cluster \\ keys}&FLOPs&AP  \\
             \midrule
             $L=24$& $\times$& 163.77 & 0.413 \\
             $\times$& $L=8$& 163.48 & 0.414 \\
             $L=32$& $L=12$& 162.7 & 0.411 \\
             \bottomrule
        \end{tabular}%}
        \caption{The AP and FLOPs under different clustering targets. We perform experiments on clustering queries, clustering keys, and clustering both queries and keys respectively.}
        \label{tab:cluster}
    \end{minipage}
    \hspace{0.2in}
    \begin{minipage}[]{0.48\textwidth}
        \centering
        \renewcommand\tabcolsep{6.0pt}
        %\scalebox{0.9}[0.85]{
        \begin{tabular}{ccc}
            \toprule
             Model & \makecell{Inference Time \\ per Image} & Memory \\
             \midrule
             DETR-DC5 & 0.246s & 1862MiB\\
             ACT(L=32) & 0.218s & 1733MiB\\
             ACT(L=24) & 0.207s & 1584MiB\\
             ACT(L=20) & 0.195s & 1415MiB\\
             ACT(L=16) & 0.183s & 1142MiB\\
             \bottomrule
        \end{tabular}%}
        \caption{The inference time and memory cost on an Nvidia GeForce GTX TITAN X with batch size of 1.}
        \label{tab:time}
    \end{minipage}}
\end{table}

\subsection{Final Performance}
\textbf{Speed Accuracy Trade-off}. In this section, we start by comparing the AP of our model with DETR-DC5. We also compare our adaptive clustering method with K-means clustering used by Vyas~\etal~\cite{vyas2020fast}. We refer to the number of clusters in K-means as $C$. We adjust $L$ and $C$ and calculate the AP under different computational budgets. As we can see in Figure~\ref{fig:results}, under an equalized computational budget, the AP of our ACT model is much higher than K-means's. Compared with DETR-DC5, we reduce the FLOPs from 184.1 GFLOPs to 168.9 GFLOPs while the loss in AP is only 0.7\%. The yellow line also shows the result of a Multi-Task Knowledge Distillation (MTKD). MTKD can significantly improve AP, especially for those models with smaller flops. Through MTKD, the AP of ACT with $L=16$ is increased by 4.3\%, and ACT with $L=32$ achieves almost the same performance as DETR-DC5.  

We also analyze the advantages of our method compared to the K-means clustering. K-means clustering uses the same number of clusters in all of the encoder layers. But it is hard the determine an appropriate hyper-parameter because the distribution of queries varies greatly with the input image and the index of the encoder layer. An inappropriate hyper-parameter $C$ will lead to bad estimates or a waste of computing resources. Another disadvantage is that K-means may not converge well in some cases, and many clusters will be empty when $C$ is relatively large. Our method adaptively determines the number of prototypes so that these disadvantages can be avoided.

\textbf{Compare AP in Detail}. We refer to the bbox AP of the large, medium, and large size instances as AP$_L$, AP$_M$, and AP$_S$ respectively. In this section, we will compare these metrics with Faster RCNN~\cite{ren2015faster} and DETR~\cite{carion2020end}. DETR and Faster RCNN use dilated ResNet-50 as backbone. Table~\ref{tab:ap} shows the results in detail. 

Our ACT model with L equal to 32 achieves the similar performance as Faster RCNN-DC5 with much fewer FLOPs. What's more, ACT is much stronger than Faster RCNN-DC5 in detecting objects with large or medium size. %Through a few-epoch of MTKD, the advantages of ACT are more obvious. 
Our ACT model can approximate the attention map in DETR well especially for large-size objects. For example, when L=32, the AP$_L$ of our model is the same as DETR-DC5. Most of the loss in AP occurs in small and medium-sized objects. Through a few-epoch of MTKD, the APs in different sizes have been significantly improved, especially for those models with fewer FLOPs. And our ACT with L equal to 32 achieves almost the same performance as DETR-DC5. %We found that most of the improvement in AP comes from small and medium-sized objects.
An interesting finding is that AP$_L$ of our ACT is 0.3\% higher than DETR's. We believe that in the process of knowledge distillation, our clustering attention can be regarded as a dropout operation, which can prevent overfitting. 
% Compared with Deformable DETR, our ACT can achieve comparable performance with fewer FLOPs. When L=32, ACT saves 4.1 GFLOPs computational cost and led by 3.1\% in AP$_L$. The AP$_M$ of ACT is only 0.3\% lower than that of Deformable DETR, and this gap can be made up by MTKD. Deformable DETR has a larger lead in AP$_S$, but most of the gain is brought by FPN.

%\textbf{Composition of FLOPs}. We also analyze the composition of FLOPs in DETR-DC5. Table~\ref{tab:flops_proportion} shows the results. As we can see, the backbone accounts for more than 60\% of the FLOPs. Then there is the attention module in the encoder, which accounts for 24.6\%. Our method focuses on reducing FLOPs in this part. For example, when L=24, we reduce the FLOPs of the attention module in the encoder from 43.5 GFLOPs to 23.2 GFLOPs. In this way, the FLOPS of attention is balanced with the FLOPS of FFN. The statistics also indicate that if we want to further reduce the FLOPs, we need to do more research on the backbone and FFN.

\textbf{Inference Time and Memory}. The above analyses are based on theoretical computation cost (FLOPs). We also tested the time and memory cost in a real environment. Table~\ref{tab:time} shows the inference time and memory cost on an Nvidia GeForce GTX TITAN X with the batch size of 1. As we can see, the acceleration of ACT in a real environment is consistent with the theoretical analysis, and the memory cost is also significantly reduced.

% \begin{table} 
%     \centering
%     \caption{The composition of FLOPs in DETR-DC5.}
%     \begin{tabular}{cccc}
%         \toprule
%          Stage&Module Type&GFLOPs&Proportion \\
%          \midrule
%          Backbone&ResNet-50 &110.7&60.1\% \\
%          Encoder&Attention &43.5&24.6\% \\
%          Encoder&FFN &22.0&12.0\% \\
%          Decoder&Attention+FFN&6.05&3.29\% \\
%          Prediction&FFN &1.85&0.01\% \\
%          \midrule
%          Total &&184.1&100\%\\
%          \bottomrule
%     \end{tabular}
%     \label{tab:flops_proportion}
% \end{table}

\section{Visualisation of Adaptive Clustering}
\begin{figure}
    \centering
    \scalebox{0.9}[0.8]{
    \begin{minipage}[t]{0.48\textwidth}
        \centering
        \includegraphics[width=5cm]{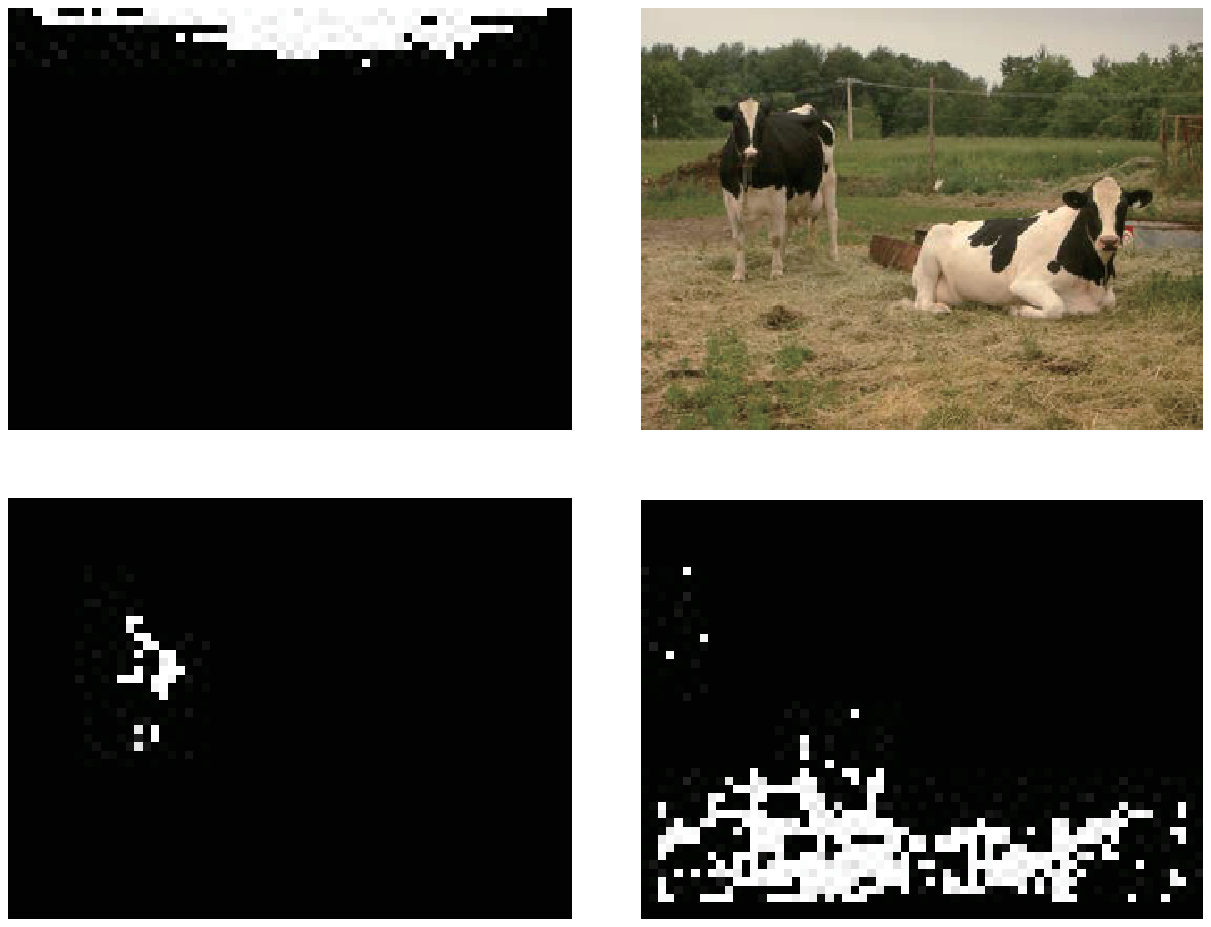}
        \caption{We visualize some representative clusters in the encoder. The queries where the white pixel is located belong to the same cluster. 
        %The original image is at the top right. The cluster on the upper left contains features of the sky, the cluster on the bottom left contains features of the cow, and the cluster on the bottom right contains features of the field.
        }
        \label{fig:cluster}
    \end{minipage}
    \hspace{0.2in}
    \begin{minipage}[t]{0.48\textwidth}
        \centering
        \includegraphics[width=5cm]{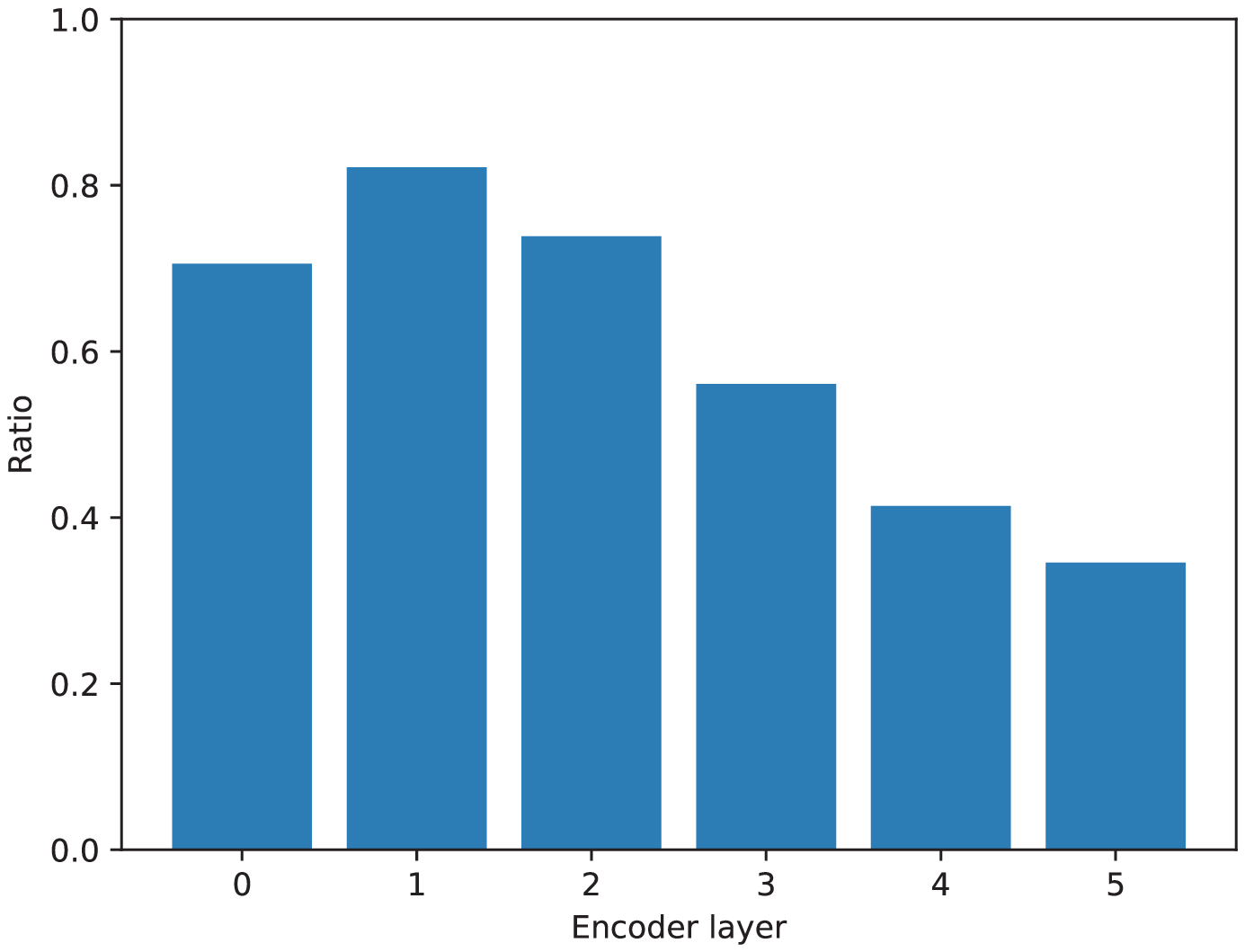}
        \caption{The ratio of the number of prototypes to the number of queries in each encoder layer. We show the average value on the first 100 images in the validation set.}
        \label{fig:prototypes}
    \end{minipage}}
\end{figure}

To analyze which queries are represented by the same prototype, we visualize some representative clusters in Figure~\ref{fig:cluster}. We can easily find that the three clusters displayed are the features of the cow, the sky, and the field. This indicates that our clustering is related to semantics and location. Those queries with similar semantics and similar locations will easily be grouped. 

To prove that our method can adaptively determine the number of prototypes based on the distribution of queries, Figure~\ref{fig:prototypes} counted the ratio of the number of prototypes to the number of queries in each encoder layer for the images on the validation set. We can find that as the encoder layer goes deeper, the number of prototypes shows a downward trend, because the features are more redundant there. This shows that our adaptive clustering method is very effective for this situation where the query distribution changes greatly.

\section{Conclusion}
% \subsection{Language}
In this paper, we propose the Adaptive Clustering Transformer (ACT) to reduce the computation and memory costs for object detection. Most previous efficient transformers need a re-training when applied to DETR. However, training the DETR requires 500 epochs approximately 1920 GPU hours for a single V100 GPU. Our proposed ACT does not need any re-training process due to the compatibility between ACT and Transformer. ACT reduces the redundancy in the pre-trained DETR in a clever adaptive clustering way. In the future, we will look into the ACT on training from scratch setting and also apply ACT to perform cross-scale information fusion over multi-scale Feature Pyramid Network (FPN)~\cite{lin2017feature}. 

\noindent\textbf{Acknowledgement.} This project is supported by National Natural Science Foundation of China---Youth Science Fund (No.62006006) and Shanghai
Committee of Science and Technology, China (Grant No.
21DZ1100100 and 20DZ1100800).

\bibliography{egbib}
\end{document}